\documentclass[10pt,twocolumn,letterpaper]{article}

\usepackage{pgfplots}

\usepackage{cvpr}
\usepackage{times}
\usepackage{epsfig}
\usepackage{graphicx}
\usepackage{amsmath}
\usepackage{amssymb}

\usepackage{color}
\definecolor{fsublue}{RGB}{62,106,190}
\definecolor{craneblue}{RGB}{4,6,76}
\usepackage{subfig}

\usepackage{footmisc}
\usepackage{booktabs}

\usepackage[pagebackref=true,breaklinks=true,letterpaper=true,colorlinks,bookmarks=false]{hyperref}

\cvprfinalcopy 

\def\cvprPaperID{????} 

\newcommand\myparagraph[1]{\paragraph{#1}}

\ifcvprfinal\fi

\begin{document}

\title{Fine-grained Recognition Datasets for Biodiversity Analysis}

\author{Erik Rodner$^1$, Marcel Simon$^1$, Gunnar Brehm$^3$, Stephanie Pietsch$^4$\\J. Wolfgang W{\"a}gele$^4$, Joachim Denzler$^{1,2}$\\[-2pt]
\small{$^1$Computer Vision Group, $^2$Michael Stifel Center Jena, $^3$Phyletisches Museum}\\[-3pt]
\small{Friedrich Schiller University Jena, Germany}\\[-3pt]
\small{$^4$Zoological Research Museum Alexander Koenig, Bonn, Germany}\\[-3pt]
\small{Project website and datasets available at: \url{http://www.inf-cv.uni-jena.de/fgvcbiodiv}}
}

\maketitle

\begin{abstract}
    In the following paper, we present and discuss challenging applications for fine-grained visual classification (FGVC): biodiversity and species analysis.
We not only give details about two challenging new datasets suitable for computer vision research with up to 675 highly similar classes, but
also present first results with localized features using convolutional neural networks (CNN).
We conclude with a list of challenging new research directions in the area of visual classification for biodiversity research.
\end{abstract}

\section{Introduction}

Fine-grained visual recognition of birds and animals has come already a long way in the last years, starting from $10\%$ recognition rate on the CUB200-2011 bird dataset in 2011~\cite{wah} to $85\%$ recently achieved by \cite{branson}.
Despite its obvious use as a benchmark for computer vision techniques, we argue that there is indeed a huge application potential for these approaches in the area of biodiversity research. 

Currently, visual recognition techniques or even image analysis tools are rarely used by biologists, although an enormous amount of expert annotation is required to build large image datasets such
as the ones of \cite{brehm} and \cite{janzen}. These datasets provide examples of highly diverse but poorly known tropical insect communities, which represent an important fraction of global biodiversity and which are functionally important in complex and endangered forest ecosystems. Furthermore, the datasets are important for understanding the changes of species composition in ecosystems caused by climate change and deforestation. 
Even when the majority of species are still unknown (as typical for tropical forests), visual discrimination allows inventorying for the goals of conservation biology.
Therefore, there is a need for automated vision systems which are able to assist experts with discrimination and annotation as well as with systematic and quantitative analysis of species
differences.

Interestingly, the expert-labeled datasets of \cite{brehm} and \cite{janzen} show that issues remain in fine-grained recognition which might have been underestimated by computer vision researchers; such as the lack of large-scale training data or 
detailed annotations as well as the need for approaches providing plausible models and visual features that can be interpreted by biologists and other experts. 
While we are briefly discussing several of these challenges at the end of the paper, 
we first introduce the datasets of \cite{brehm} and \cite{janzen}, which we prepared for FGVC research, as well as results we were able to obtain with current techniques.

    \begin{table*}[tb]
    \centering
    \begin{tabular}{lccccc}
    \toprule
    dataset & \#classes & \#images & \#images for training & accuracy (global) & accuracy (pyramid)\\
    \midrule
    Ecuador moth dataset~\cite{brehm}& 675 & 2120 & 1445 & $55.7\%$ & $53.5\%$\\
    Costa rica dataset~\cite{janzen}& 331 & 3224 & 992 & $79.5\%$ & $82.1\%$ \\
    \bottomrule
    \end{tabular}
    \caption{Categorization results for the two biodiversity datasets (butterflies and moths) of \cite{brehm} and \cite{janzen}.}
    \label{tab:datasets}
    \end{table*}
    
    \begin{figure*}[tb]
        \includegraphics[height=0.122\linewidth]{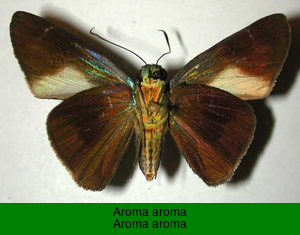}
        \includegraphics[height=0.122\linewidth]{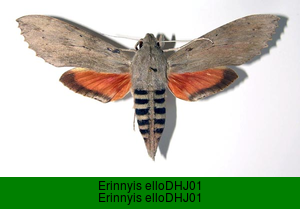}
        \includegraphics[height=0.122\linewidth]{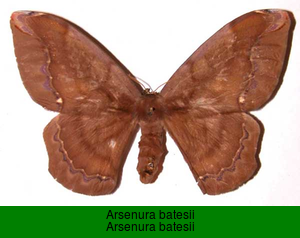}
        \includegraphics[height=0.122\linewidth]{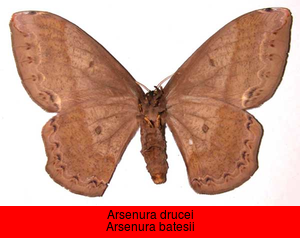}
        \includegraphics[height=0.122\linewidth]{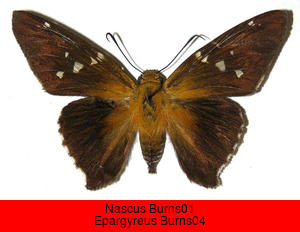}
        \includegraphics[height=0.122\linewidth]{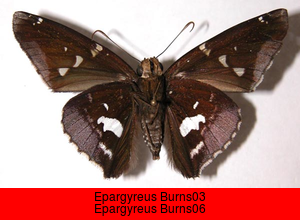}
        \caption{Example classification results for the Costa Rica dataset (input image, predicted label, ground-truth label). Images are directly obtained from \cite{janzen} and have the following identifiers: 
            \small{00-SRNP-1311-DHJ33001, 00-SRNP-1536-DHJ95316, 00-SRNP-4253-DHJ36384, 00-SRNP-4253-DHJ36385, 01-SRNP-16434-DHJ305668, 03-SRNP-20073-DHJ91439}.}
        \label{fig:qualitative}
    \end{figure*}

\section{New FGVC biodiversity datasets}
\label{sec:datasets}

    \begin{figure*}
        \centering
        \includegraphics[width=0.9\linewidth]{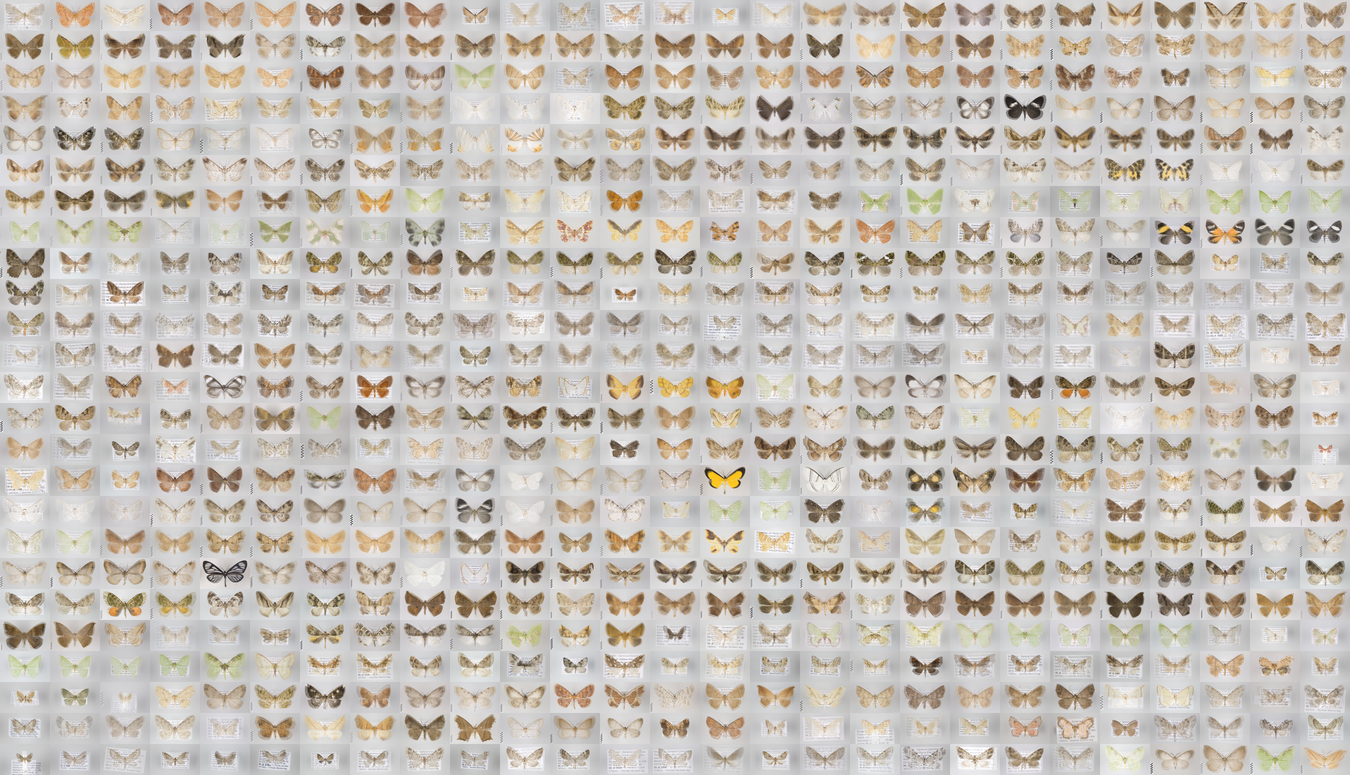}
        \includegraphics[width=0.9\linewidth]{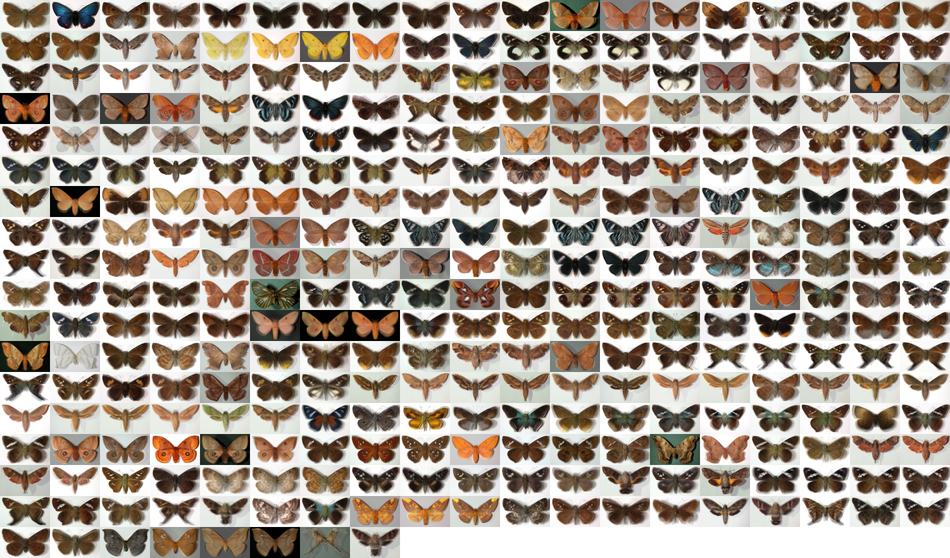}
        \caption{Average images of all categories of the Ecuador and the Costa Rica dataset.}
        \label{fig:datasets}
    \end{figure*}

    In the following, we present two datasets (\figurename~\ref{fig:datasets}), which are ready to use for computer vision researchers. All images show moths and butterflies with artificially spread wings. While uncommon in natural photos, this is the way animals are 
    prepared for scientific collections to expose the features of the hind wings, which normally are covered by the anterior wings in living specimens.
    In both datasets, species sorting was achieved by a combination of traditional sorting by specialists, according to external characters, and the use of so called DNA barcoding, \ie the use of a standardized gene fragment of the mitochondrial gene which allows delineating species even in difficult, cryptic, and small taxa~\cite{barcoding}. 

    \myparagraph{Ecuador moth dataset~\cite{brehm}}

        The dataset of \cite{brehm} includes only one single family of moths (Geometridae) quantitatively collected in montane tropical rainforests in southern Ecuador, 
        the global diversity hotspot of this taxon. Our dataset covers $675$ observed and genetically verified species in the area. It includes many closely related and look-alike species, 
        most of them unknown to science, and is therefore particularly challenging. Since expert knowledge on these moths is very scarce, automated image analysis could substantially contribute to 
        species-sorting by untrained persons, or to monitoring schemes in endangered habitats.
        The images have been taken in a controlled environment with uniform background and canonical poses, which makes it easy to focus feature extraction on the important
        parts of the image. Since the dataset only includes a few images per species, we use male and female individuals within one category. We will release a challenging subset of the dataset to the public.

    \myparagraph{Costa rica dataset~\cite{janzen}}
        The dataset of \cite{janzen}, derived from long-term sampling and caterpillar rearing, includes a broad range of moth and butterfly taxa sampled in north western Costa Rica.
        Since we have a larger initial dataset, we reduced it to female individuals only and species with at least 2 images. 
        The dataset is already publicly available and we plan to release converted meta data and links to ease
        its use for the computer vision community. Furthermore, a large part of it is already linked in the encyclopedia of life database~\footnote{\url{http://eol.org}},
        where additional meta information is likely to be published in future.

\section{Global and pyramid-based CNN baseline}

    How well do current vision technologies perform on the datasets presented?
    Since most of the animals in the images of both datasets are already aligned, we computed global CNN features with AlexNet~\cite{alexnet} (caffe reference network) using layer \texttt{pool5} and
    used a one-vs-all linear SVM for classification. For the Ecuador dataset~\cite{brehm}, all except of one randomly selected image for each category
    has been used for training. Learning on the Costa Rica dataset was done with up to three training examples for each category.

    Table~\ref{tab:datasets} gives the accuracies for each of the datasets. At a first glance, although the number of classes is extremely high,
    we are able to achieve reasonable accuracies. The dataset is far more challenging than the Leeds butterfly dataset of \cite{leedsdataset} with $10$ categories,
    where we are able to obtain an accuracy of $99.24\%$ with the same techniques.

    To focus on more subtle differences in just a few parts (different colors of parts of the wing for example), 
    we calculate a spatial pyramid with two levels using CNN features.
    First, global features for the whole image are calculated. 
    Then the image is divided into four equal-sized subregions and all features are concatenated.
    The spatial pyramid helps to improve the accuracy by $2.6\%$ for the Costa Rica but not for the Ecuador dataset (Table~\ref{tab:datasets}). \textit{Please note that both datasets contain a certain dataset bias, which is discussed in more detail on the project website (see header)}.

\section{Conclusions and upcoming challenges} 

As we have seen in the brief description of our first experiments, vision algorithms can already obtain a suitable accuracy for challenging species identification tasks. However,
automated classification is not the only research direction in the area of computer-assisted biodiversity research and we list a few upcoming challenges:
\begin{enumerate}
\itemsep-4pt
\item Open-set recognition for counting known species and automatically detecting novel ones: biologists and citizen scientists need tools that allow them to detect animals that are likely going to belong to a new species. This would allow
    for a certain pre-filtering of animals prior to comprehensive DNA barcoding analysis. Furthermore, it could be also used to derive quantitative measures for biodiversity research~\cite{Bodesheim15:LND}.
\item Incorporating human-machine interaction not only for active classification~\cite{wah11-mrp} and learning~\cite{alex}: There is a lot of expert knowledge already
    available which should be used to develop new models or actively guide the search for relevant features during learning.
\item Discovering interpretable features: automatically relating learned models to human-interpretable features would enable biologists to study especially hard to differentiate
species in more detail. 
\item Dealing with only a few training examples~\cite{Rodner10:OSL}: we need to build fine-grained recognition systems, which are especially able to deal with rare classes. This is important since currently available
and important biodiversity datasets (see Section~\ref{sec:datasets}) are mostly comprised of classes with only up to 5 training examples.
\item Deriving compact textual and discriminative descriptions of the visual differences between the species.
\end{enumerate}

\noindent
{\small
\bibliographystyle{plain}
\bibliography{paper}
}

\end{document}